\documentclass[letterpaper]{article} 
\usepackage{aaai2026}
\usepackage{times}  
\usepackage{helvet}  
\usepackage{courier}  
\usepackage[hyphens]{url}  
\usepackage{graphicx} 
\usepackage{natbib}  
\usepackage{caption} 
\usepackage{algorithm}
\usepackage{algorithmic}
\usepackage{float}
\usepackage{multirow}

\usepackage{newfloat}
\usepackage{listings}
\DeclareCaptionStyle{ruled}{labelfont=normalfont,labelsep=colon,strut=off} 
\floatstyle{ruled}
\newfloat{listing}{tb}{lst}{}
\floatname{listing}{Listing}
\title{4DGS-Craft: Consistent and Interactive 4D Gaussian Splatting Editing}
\author{
    Lei Liu$^1$, Can Wang$^1$, Zhenghao Chen$^2$, Dong Xu$^1$
}
\affiliations{
    \textsuperscript{\rm 1}The University of Hong Kong, Hong Kong SAR, China \\
    \textsuperscript{\rm 2}The University of Newcastle, Newcastle, Australia \\ 
}

\usepackage{bibentry}
\usepackage{amsmath}
\usepackage{xspace}
\newcommand{\method}{\textit{4DGS-Craft}\xspace}

\nocopyright

\begin{document}

\maketitle






\begin{abstract}
Recent advances in 4D Gaussian Splatting (4DGS) editing still face challenges with view, temporal, and non-editing region consistency, as well as with handling complex text instructions. To address these issues, we propose \method, a consistent and interactive 4DGS editing framework.
We first introduce a 4D-aware InstructPix2Pix model to ensure both view and temporal consistency. This model incorporates 4D VGGT geometry features extracted from the initial scene, enabling it to capture underlying 4D geometric structures during editing. We further enhance this model with a multi-view grid module that enforces consistency by iteratively refining multi-view input images while jointly optimizing the underlying 4D scene. Furthermore, we preserve the consistency of non-edited regions through a novel Gaussian selection mechanism, which identifies and optimizes only the Gaussians within the edited regions.
Beyond consistency, facilitating user interaction is also crucial for effective 4DGS editing. 
Therefore, we design an LLM-based module for user intent understanding. This module employs a user instruction template to define atomic editing operations and leverages an LLM for reasoning. As a result, our framework can interpret user intent and decompose complex instructions into a logical sequence of atomic operations, enabling it to handle intricate user commands and further enhance editing performance.
Compared to related works, our approach enables more consistent and controllable 4D scene editing. Our code will be made available upon acceptance.
\end{abstract}

\section{1. Introduction}
\begin{figure*}[htbp]
    \centering
    \includegraphics[width=\linewidth]{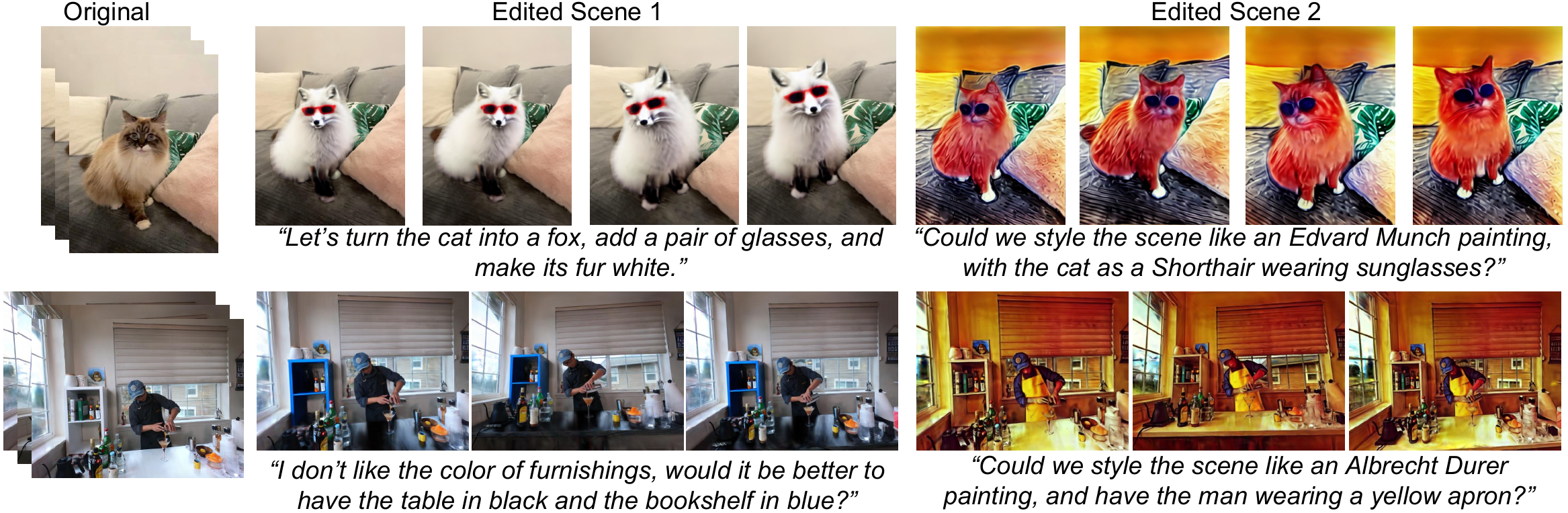}
    \vspace{-6mm}
    \caption{We present \method, an intuitive and consistent 4D scene editing framework that accommodates a wide range of user instructions, including highly abstract and complex commands, and delivers visually pleasing results.}
    \vspace{-5mm}
    \label{fig:teaser}
\end{figure*}



The emergence of advanced 3D representations, such as Neural Radiance Fields (NeRF)~\cite{mildenhall2021nerf} and 3D Gaussian Splatting (3DGS)~\cite{kerbl20233d}, has transformed real-world scene acquisition and enabled photorealistic novel-view synthesis. However, these methods are limited to modeling static scenes. To capture dynamic environments, recent works have extended NeRF and 3DGS by incorporating a temporal dimension, facilitating 4D scene reconstruction through techniques such as deformation fields~\cite{park2021nerfies,pumarola2021d,wu20244d}. As Gaussian Splatting provides faster reconstruction and rendering through explicit point clouds of Gaussian primitives and a differentiable rasterizer, there is a growing demand for editing 4DGS representations~\cite{shao2024control4d,kwon2025efficient,he2025ctrl}.

Recent 4DGS editing methods leverage 2D pre-trained diffusion models, particularly InstructPix2Pix (IP2P)~\cite{brooks2023instructpix2pix}, to guide the optimization process for 4DGS editing~\cite{he2025ctrl}. Benefiting from the generative capabilities of 2D diffusion models, these approaches can easily produce natural and realistic editing results. However, 2D diffusion models struggle to provide view and temporally consistent supervision, often resulting in geometry and texture inconsistencies in both edited and non-edited regions.
Beyond consistency issues, current 4DGS editing methods face two fundamental limitations: inadequate incorporation of user editing intent and restriction to oversimplified text instructions. These constraints significantly hinder their practical utility in real-world applications.

In this paper, we propose \method, a framework that significantly improves consistency in 4DGS editing and enables understanding of user intent to support complex and abstract instructions. 
Our approach is motivated by four key insights. 1) First, recent feedforward 3D attribute estimation models, such as VGGT~\cite{wang2025vggt}, provide a robust backbone for learning 3D-consistent geometry feature representations. This raises an important question: \textit{can incorporating VGGT features help promote view and temporal geometry consistency in 4DGS editing?} 2) Second, since single-view IP2P cannot provide consistent supervision during the optimization process, a question arises: \textit{can we design a multi-view IP2P to improve consistency?} 3) Third, in GS editing, Gaussian primitives may be cloned, split or pruned. \textit{Given only a single-frame editing mask, can we track which Gaussians should be optimized in conjunction with these changes, in order to preserve non-edited regions unchanged?} 4) Lastly, although user-provided editing prompts can span a broad spectrum—from abstract to concrete and from complex to simple, current methods remain limited to handling only single, unambiguous text instructions of limited complexity. Inspired by recent advances in large language models (LLMs) for reasoning~\cite{wang2024chat2layout,zhao2023large}, \textit{could we develop an LLM-based module specialized for editing, to better understand user intent and enhance editing performance?}

Our method addresses the above questions through the following technical contributions:

\begin{itemize}
\item We propose a 4D-aware IP2P network that first enhances IP2P by integrating 4D VGGT features extracted from the initial scene, enabling the generation of geometrically consistent 2D supervision signals for iterative 4D scene optimization. Furthermore, this network augments IP2P with a multi-view grid module that iteratively incorporates multi-view renderings during the optimization process, allowing for multi-view-aware scene updates and improved consistency throughout optimization.
\item We introduce a novel Gaussian selection mechanism that robustly tracks 3D Gaussian primitives across both edited and non-edited regions, maintaining consistency even when primitives undergo cloning, splitting, or pruning operations. This enables easy supervision to ensure that the non-editing region remains unchanged.
\item We design an LLM-based intent understanding module that employs structured instruction templates to define atomic editing operations, leveraging the LLM’s reasoning capabilities to decompose complex and abstract user instructions into simple, concrete sequences of editing operations. This approach enables our framework to interpret sophisticated instructions and significantly enhances overall editing performance.
\item We demonstrate that our \method enables intuitive and consistent 4D scene editing by effectively handling a wide range of user instructions, including highly abstract and complex commands, while achieving visually pleasing results (see Figure~\ref{fig:teaser}) and outperforming baselines on various metrics.
\end{itemize}
\section{2. Related Work}

\noindent \textbf{Static NeRF and 3DGS Editing.}
The photo-realistic rendering capabilities of NeRF and GS have attracted significant research interest in editing both representations. Early works attempt to edit NeRF or GS by optimizing the representations with CLIP guidance ~\cite{wang2022clip,wang2023nerf} or SDS distillation~\cite{pooledreamfusion}. CLIP guidance offers a semantic editing direction but struggles with fine-grained control, while SDS provides more diverse supervision but can easily lead to oversaturation issues~\cite{wang2023prolificdreamer}. Instruct-NeRF2NeRF~\cite{haque2023instruct} presents a different solution by lifting 2D supervision to NeRF; it uses IP2P to iteratively edit the rendered views from the scene as supervision during the optimization process, enabling more diverse and realistic editing results. More recently, some works have explored articulating and deforming the geometry of NeRF and GS using meshes~\cite{wang2023mesh,gao2025mani}, point clouds~\cite{chen2023neuraleditor}, or cages~\cite{peng2022cagenerf,jambon2023nerfshop,jiang2024vr}. However, these methods are designed for static scenes. In contrast, our work targets dynamic scenes, which are more challenging due to the added temporal dimension.



\noindent \textbf{Dynamic NeRF and 4DGS Editing.}
Recent advancements have extended NeRF and GS to dynamic representations by modeling both spatial and temporal dimensions. Many approaches define a canonical space to represent static parts of a scene and introduce a deformation field~\cite{park2021nerfies,pumarola2021d,wu20244d,yangreal} to capture time-aware motion variations within the 3D scene.
Similar to 3D editing, recent 4D editing works have also explored using IP2P to edit dynamic NeRF and 4DGS.
For example, Instruct 4D-to-4D~\cite{mou2024instruct} enhances IP2P with an anchor-aware attention module and integrates optical flow-guided appearance propagation to enable consistent editing of dynamic NeRFs.
Inspired by Instruct 4D-to-4D, CTRL-D~\cite{he2025ctrl} uses IP2P to edit 4DGS. It first edits a single frame as a reference to fine-tune IP2P for producing personalized images. Then, it iteratively applies IP2P to supervise scene optimization, achieving consistent 4D editing.
Instruct-4D further employs Coherent-IP2P~\cite{wu2023tune}, which replaces the 2D convolutional layer with a 3D convolutional layer to improve spatial consistency. However, we found that these methods still suffer from inconsistent editing, as the IP2P they use does not fully account for geometric and texture consistency both spatially and temporally. In contrast, we propose a 4D-aware IP2P that incorporates 4D VGGT~\cite{wang2025vggt} features and introduces a multi-view grid module to enable more consistent editing.

\begin{figure*}[h]
    \centering
    \includegraphics[width=\linewidth]{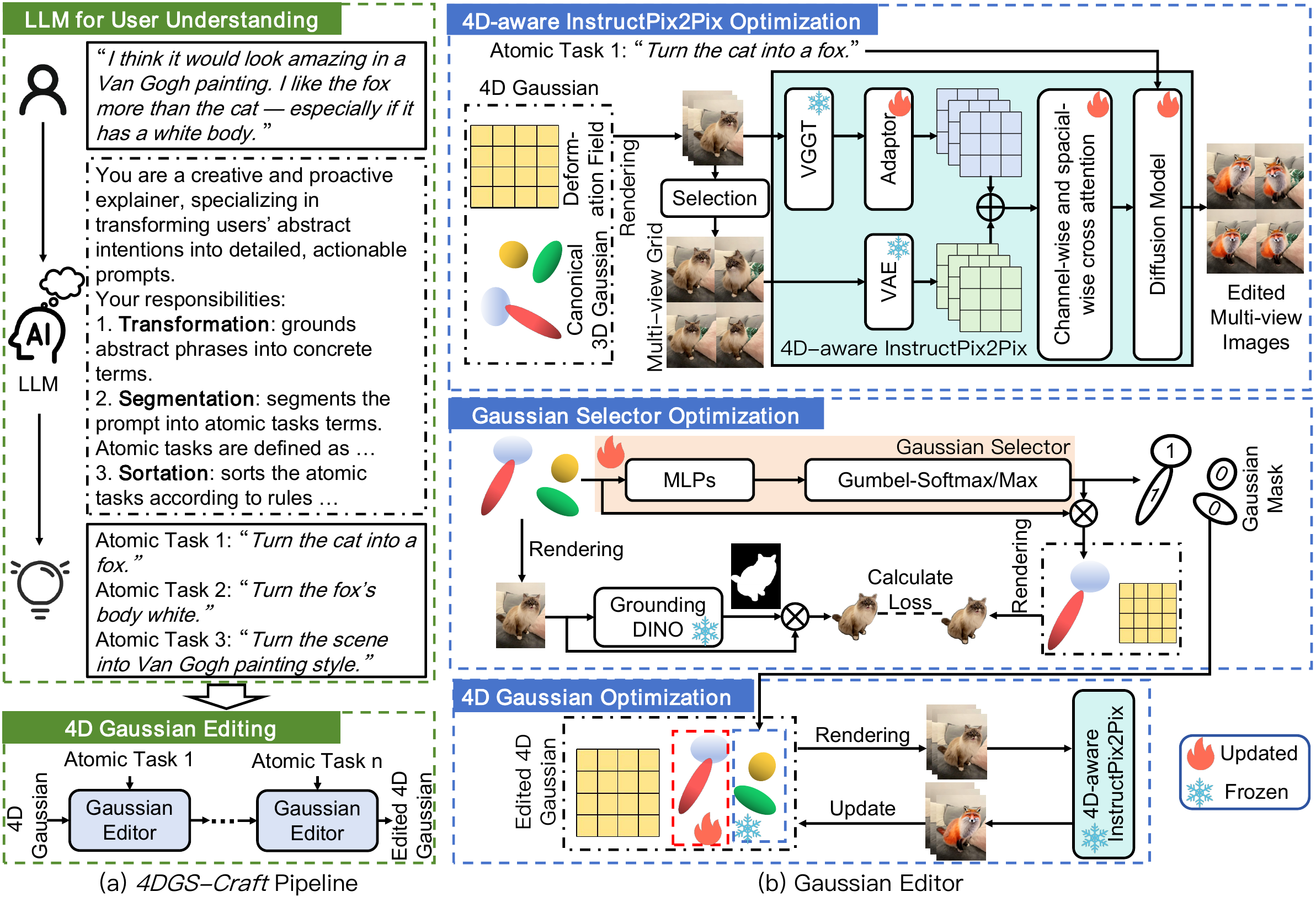}
    \vspace{-5mm}
    \caption{\textbf{Framework.} (a) \method Pipeline, which first employs the LLM-based Intent Understanding Module to decompose the editing prompt into atomic editing operations and subsequently applies the Gaussian Editor to edit the 4DGS based on those atomic tasks. (b) Details of the Gaussian Editor, which performs 4DGS editing in three stages: 4D-aware InstructPix2Pix optimization, Gaussian Selector optimization, and 4D Gaussian optimization.}
    \label{fig:framework}
    \vspace{-5mm}
\end{figure*}

\noindent \textbf{LLM Reasoning.}
Large Language Models (LLMs) have proven to be effective tools for user understanding and task planning~\cite{wang2024chat2layout,zhao2023large}. For example, Chat2Layout~\cite{wang2024chat2layout} defines a set of atomic operations for 3D layout synthesis and leverages LLMs to decompose open-set instructions into sequences of these atomic tasks, thereby supporting a broad and flexible vocabulary. 
Inspired by this method, we have designed an LLM module to understand user intent in 4DGS editing. Rather than directly using the LLM to decompose a complex user prompt into a sequence of simpler ones, our LLM module defines a set of atomic operations specific to 4DGS editing, and then decomposes complex instructions into sequences according to specific rules. This module can also interpret abstract instructions and convert them into structured, concrete representations. We demonstrate that our LLM module enables a better understanding of open-set user instructions and facilitates improved editing performance in a chain-of-thought (CoT) manner.

\section{3. Method}

The pipeline of our method is illustrated in Figure~\ref{fig:framework}. Given an initial reconstructed dynamic scene represented by 4DGS (Section 3.1) and a user prompt, our method first utilizes an LLM to interpret user intent, converting complex or abstract textual instructions into a sequence of atomic editing tasks described in text (Section 3.2). For each atomic task, we edit a single frame of the 4D scene based on the text description, using it as a reference to fine-tune a 4D-aware IP2P network. We then leverage this 4D-aware IP2P network to supervise the iterative scene optimization process and achieve the desired editing effects (Section 3.3). To preserve non-edited regions, we train a Gaussian selector that identifies which Gaussian primitives should be edited. During optimization, we track the Gaussian primitives in non-edited regions and keep them unchanged (Section 3.4). Our framework iteratively processes each atomic task to obtain the final result.

\subsection{3.1 Dynamic Scene Reconstruction}

We reconstruct a dynamic scene with camera annotations using 4DGS, which consists of a canonical 3D Gaussian $G$ to represent the static components and a deformation field $F_{deform}$ to capture the motions.
Here, $G=\{g_1, \dots, g_N\}$ consists of $N$ Gaussian primitives, where each Gaussian $g_i=(p_i,s_i,r_i,o_i,c_i)$. $g_i$ is defined by its position $p_i$, scaling vector $s_i$, rotation quaternion $r_i$, opacity $o_i$, and color parameter $c_i$. The deformation network $F_{deform}$ aims to predict an offset for the Gaussian parameters at any given time $t$: $\Delta g_i=F_{deform}(g_i, t)$. The final 4DGS can be represented as $G=(g_1+\Delta g_1, \dots, g_N+\Delta g_N)$, and can be rendered into images using standard 3DGS rasterization.
We will perform the editing on this pre-trained 4DGS.


\subsection{3.2 LLM for User Understanding}

Text-guided 4DGS editing must handle diverse user instructions, from simple to complex and from abstract to concrete, but current methods support only simple inputs. We thus propose leveraging an LLM to interpret richer user intents. 

We begin by surveying user requirements and formulating a set of atomic tasks—fundamental operations that enable complex edits. Below is our taxonomy of these tasks: 1) Color Adjustment: \textit{e.g.}, ``\textit{Repaint the wall blue}'';
2) Texture Replacement: \textit{e.g.}, ``\textit{Replace wooden flooring with marble}'';
3) Material Properties: \textit{e.g.}, ``\textit{Change from metal to wood}'';
4) Local Geometry Modification: \textit{e.g.}, ``\textit{Add a hat to the cat}''; 5) Category Swapping: \textit{e.g.}, ``\textit{Convert the dog into a cat}'';  6) Style Transfer: \textit{e.g.}, ``\textit{Change to cyberpunk style}''; 7) Background Editing: \textit{e.g.} ``\textit{Set the background to a forest}''. All these atomic tasks represent concrete, indivisible operations.

With these atomic tasks defined, we design a text prompting technique to decompose open-set user instructions into sequential atomic tasks, following a Chain-of-Thought process, as illustrated in Figure~\ref{fig:framework}. As an example, for user prompt ``\textit{I think it
would look amazing in a Van Gogh
painting. I like the fox more than the cat —
especially if it has a white body. }'', the LLM module: 1) grounds abstract phrases into concrete terms (`` \textit{Turn the sence into Van Gogh painting style. Turn the cat into a fox. Turn the fox’s body white.}''); 2) segments the prompt into a sequence of atomic tasks (1. ``\textit{Turn the sence into Van Gogh painting style}'', 2. ``\textit{Turn the fox’s body white}'', 3. ``\textit{Turn the cat into a fox}''); and 3) orders these atomic tasks based on the following criteria: Dependency Analysis—identifying prerequisite relationships between tasks, and Complexity Assessment—prioritizing tasks according to their difficulty (1. ``\textit{Turn the cat into a fox}'', 2. ``\textit{Turn the fox’s body white}'', 3. ``\textit{Turn the sence into Van Gogh painting style}''). 
By evaluating inter-task dependencies, temporal constraints, and complexity, our LLM module ensures valid editing operations.

\begin{figure*}
    \centering
    \includegraphics[width=\linewidth]{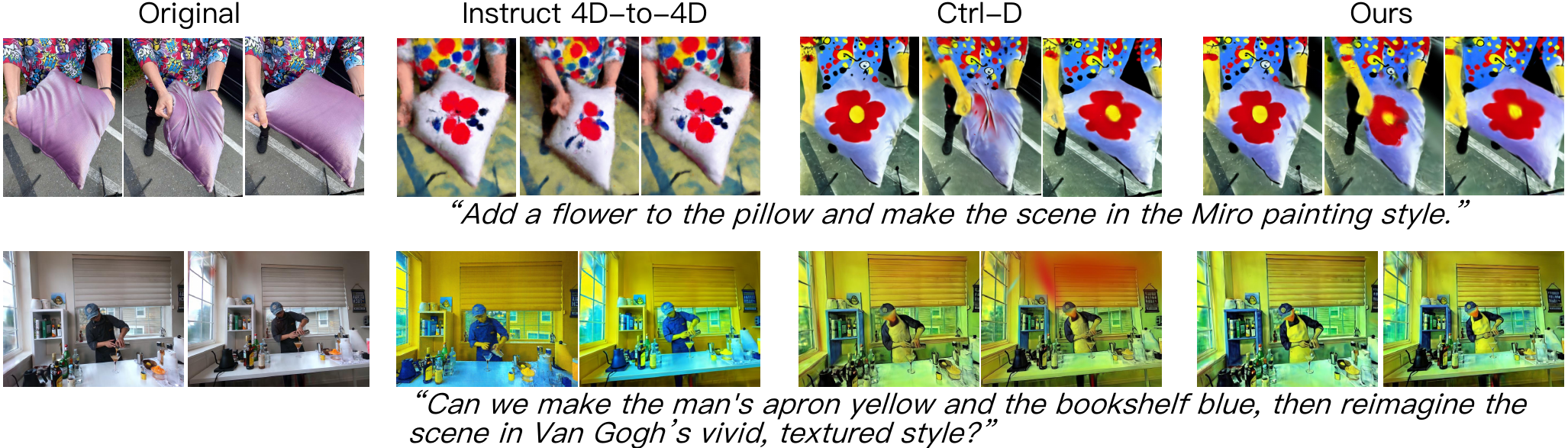}
    \vspace{-5mm}
    \caption{\textbf{Qualitative Comparison.} We provided the qualitative comparison with both the 4DGS-based editing method CTRL-D and the dynamic NeRF-based editing method Instruct 4D-to-4D. Our results demonstrate superior prompt-following capability and higher consistency across views and time.}
    \vspace{-4mm}
    \label{fig:comparable}
\end{figure*}

\subsection{3.3 4D Gaussian Editing}

Our editing process for each atomic task also follows the IP2P-based iterative optimization framework, similar to Instruct 4D-to-4D~\cite{mou2024instruct} and CTRL-D~\cite{he2025ctrl}. However, unlike these previous methods, which do not account for spatial and temporal consistency within IP2P, we propose a 4D-aware IP2P that explicitly enforces both spatial and temporal consistency during optimization.

Inspired by the recent success of VGGT~\cite{wang2025vggt} in consistent geometry inference, we incorporate 4D VGGT features into the IP2P framework. Specifically, given a multi-view temporal image sequence $I_{seq}=\left [ I_i^{} \right ]_{i=1}^{M}$ rendered from the initial 4DGS $F_{\text{4DGS}}$, we utilize a pre-trained VGGT backbone $F_{vggt}$ to extract geometry-aware features. We then design an adapter $F_{adpater}$ to transform these features, which are subsequently added as a residual to the IP2P network.
While VGGT focuses solely on geometry-consistent features and does not provide texture-aware consistency, we further enhance our approach by introducing a multi-view grid module into IP2P. This module not only strengthens geometry consistency but also incorporates multi-view texture-aware features.
We randomly sample four views $I_{grid}=[I_i^t]_{i=1}^4$ at time $t$ from $I_{seq}$
and combine them into a single image, which replaces the original source image input in IP2P. Although this is a simple design, the cross-attention mechanism within IP2P automatically computes spatial relationships across the four views, enabling the model to perceive and enforce multi-view consistency.
Finally, our 4D-aware IP2P is defined as:
\begin{equation}
    F_{\text{4D-IP2P}}=F_{\text{IP2P}}(F_{adapter}(F_{vggt}(I_{seq})),F_{vae}(I_{grid}),c)
\end{equation}
where $F_{\text{IP2P}}$ is the original IP2P network, $F_{vae}$ is the image VAE encoder, and $c$ is the text prompt.


With this 4D-aware IP2P $F_{\text{4D-IP2P}}$ prepared, given an atomic task, we first perform text-guided 2D editing on a single frame from $I_{seq}$ to obtain a reference image $I_{Edit}$. We then fine-tune $F_{\text{4D-IP2P}}$ using $I_{Edit}$ to obtain a personalized network, denoted as $F_{\text{4D-IP2P}}^*$. 
We use $F_{\text{4D-IP2P}}^*$ to provide 4DGS optimization direction in an iterative manner:
\begin{equation}
    \mathcal{L}_{opt}=\left \| F_{\text{4DGS}}(t,v) - F_\text{4D-IP2P}^*(I_{seq},I_{grid},c) \right \| _2^2
\end{equation}
where $t$ is the time and $v$ is the camera view. During optimization, we also use the Iterative Dataset Update technique in Instruct-NeRF2NeRF to update the supervision of $F_\text{4D-IP2P}^*$.

\begin{figure}[t!]
    \centering
    \includegraphics[width=\linewidth]{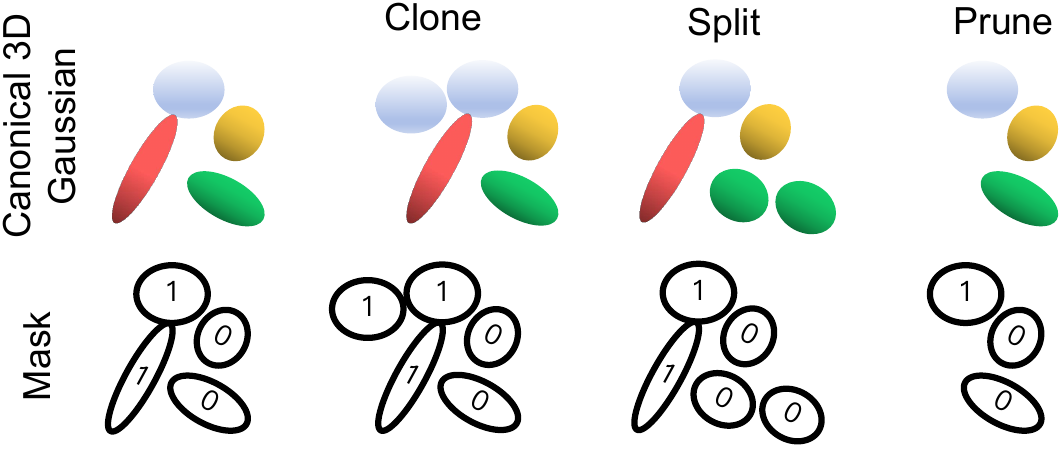}
    \vspace{-3mm}
    \caption{\textbf{Gaussian Mask Tracking.} We propose tracking Gaussian masks during the clone, split, and prune operations performed on the canonical 3D Gaussian. An indicator value of 0 denotes the non-editing region, while a value of 1 denotes the editing region. Gaussians assigned a value of 0 will not be optimized during editing.}
    \vspace{-4mm}
    \label{fig:tracking}
\end{figure}

\subsection{3.4 4D Gaussian Selection}

The proposed 4D-aware IP2P effectively preserves geometric and texture consistency within the edited regions; however, the diffusion process may inadvertently impact the consistency of non-edited regions.
To address this issue, we propose a Gaussian selection technique that localizes and tracks non-edited Gaussian primitives during the IP2P optimization process. This enables us to introduce a loss term that constrains the unedited Gaussians to remain unchanged throughout the editing process. This technique consists of two stages:

\noindent\textbf{Gaussian Mask Initialization.}
We begin by predicting the initial binary Gaussian masks $G_m$ with length $N$ where $N$ denotes the number of Gaussians in the reconstructed 4DGS.
In this mask, $G_m[i]=1$ indicates that the 
$i$-th Gaussian belongs to the editable region, while
$G_m[i]=0$ signifies membership in the non-editable region. The mask $G_m$ is obtained by training a Gaussian selector, as illustrated in Figure~\ref{fig:framework}.
To learn a binary mask, a max operation is typically required; however, this operation is non-differentiable, which poses challenges for gradient-based optimization. To address this, we employ a differentiable Gumbel-Softmax~\cite{jang2016categorical} strategy to approximate the max operation, thereby enabling gradient backpropagation during training.
We then train this Gaussian selector using a 2D segmentation loss: 
\begin{equation}
    \mathcal{L}_{mask}=\left \| \mathcal{R}(F_{\text{4DGS}}(t,v))-F_{seg}(I_{Edit},c) \cdot I_{Edit}   \right \|_2^2
\end{equation}
where $\mathcal{R}(\cdot )$ renders masked image from 4DGS $F_{\text{4DGS}}$ by filtering the Gaussian points whose mask indices are zero, and $F_{seg}$ estimates a segmentation mask 
from the editing image $I_{Edit}$ using the prompt-based segmentation method GroundDINO~\cite{liu2024grounding}.

\noindent\textbf{Gaussian Mask Tracking.}
We are then use the learned mask to indicate which Gaussians should be optimized and which should be frozen during the IP2P iterative optimization process.
However, during 4DGS optimization, Gaussian points may undergo operations such as cloning, splitting, and pruning. Thus, we propose to track Gaussian masks during optimization, as shown in Figure~\ref{fig:tracking}.
Specifically, during Gaussian point operations, the mask vector is updated accordingly. 
When a Gaussian point $g_i$ is cloned to create $g_j$, the mask vector duplicates the corresponding entry, adding
$m_j$ and setting $m_j = m_i$. When $g_i$ splits into $g_a$ and $g_b$, the mask vector adds $m_a$ and $m_b$, both initialized to $m_i$, and removes $m_i$ from the mask vector.  
If a Gaussian point $g_k$ is pruned, its associated mask entry $m_k$ is also removed. 
This design enables the mask to accurately track Gaussian points through dynamic topology changes. 

\section{4. Experiments}

\subsection{4.1 Experimental Setup}

\noindent\textbf{Dataset.} Following previous works~\cite{he2025ctrl,mou2024instruct}, we evaluate \method on the DyCheck~\cite{dycheck} and N3DV~\cite{N3DV} datasets. DyCheck is a monocular dataset composed of object-centric dynamic scenes, while N3DV is a multi-view dataset that captures dynamic indoor environments using multiple synchronized cameras.


\noindent\textbf{Implementation Details.}
Our method is implemented in Python, and all experiments are conducted on an 80GB NVIDIA A800 GPU. 
Our 4D-aware IP2P network is fine-tuned from IP2P~\cite{brooks2023instructpix2pix} with 1,000 steps. During this process, we follow the training strategy of the baseline method CTRL-D~\cite{he2025ctrl}, where the primary objective is to minimize the error between the predicted noise and the actual noise in the target-edited image.
Subsequently, the Gaussian Selector is trained for 3,000 steps using an MSE loss between the masked image and the rendering image from the masked 4DGS. Finally, we optimize the 4DGS for 20,000 steps using the standard 4DGS optimization objective~\cite{4dgs}. More details on the loss functions and training hyperparameters are stated in the supplementary materials.



\begin{figure*}[htbp]
    \centering
    \includegraphics[width=\linewidth]{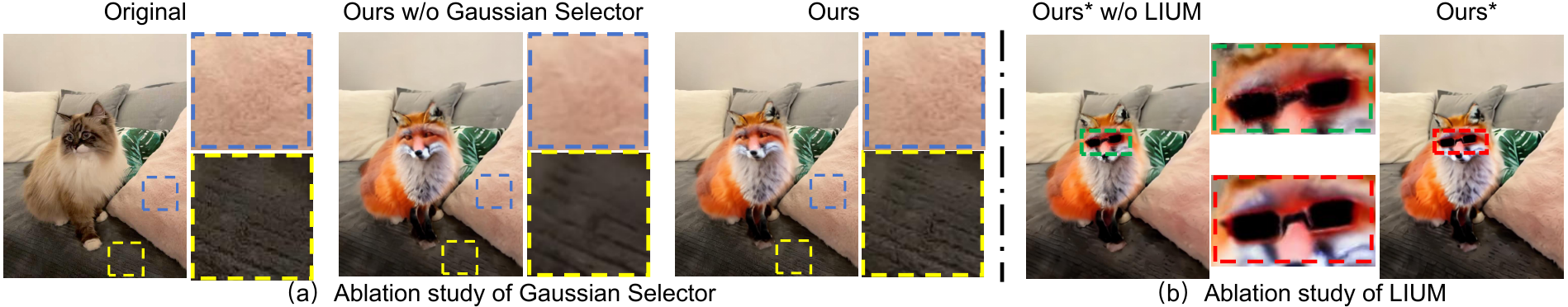}
    \vspace{-5mm}
    \caption{(a) Ablation study of Gaussian Selector on the ``\textit{mochi-high-five}'' scene from the DyCheck dataset. Our complete method demonstrates superior preservation of texture details in the non-edited area. (b) Ablation study of the LLM-based Intent Understanding Module (LIUM). $\mathrm{Ours^*}$ denotes our method without the 4D-aware IP2P and the Gaussian Selector. LIUM empowers our model to produce outputs with greater clarity and finer details in the edited area.}
    \vspace{-3mm}
    \label{fig:ab_mask_LIUM}
\end{figure*}

\subsection{4.2 Comparisons}

\noindent\textbf{Qualitative Comparison.}
The qualitative comparison results are shown in Figure~\ref{fig:comparable}, where we compare \method with CTRL-D~\cite{he2025ctrl} and Instruct 4D-to-4D~\cite{mou2024instruct}. First, we edit the ``pillow'' scene from the DyCheck dataset using the prompt ``\textit{Add a flower to the pillow and make the scene in the Miro painting style}''. Compared to CTRL-D, our method exhibits higher consistency, particularly when the pillow is folded. Second, we edit the ``coffee\_martini'' from the N3DV~\cite{N3DV} dataset with the prompt ``\textit{Make the man's apron yellow, make the bookshelf blue, and turn the scene into the Van Gogh painting style.}''. Compared to CTRL-D, our results maintain stronger consistency across views and time, especially in the regions with complex textures such as the curtain.
Compared to Instruct 4D-to-4D, our method more faithfully adheres to the given prompts—for example, in correctly editing the color of the bookshelf, apron, and table. These qualitative results demonstrate that our approach outperforms the state-of-the-art 4DGS editing method CTRL-D~\cite{he2025ctrl} as well as the dynamic NeRF-based method Instruct 4D-to-4D~\cite{mou2024instruct}, highlighting the superior editing capability of \method in 4D scene manipulation.

\begin{table}[]
    \centering
    \begin{tabular}{c|c|c|c|c}
    \hline
       Scene & Methods & $\rm{CLIP_{S}}$ & $\rm{CLIP_{D}}$ & Consistency \\ \hline
        \multirow{2}{*}{\textit{Pillow}} & Ours  & \textbf{0.192} & \textbf{0.070} & \textbf{0.942} \\
        & CTRL-D & 0.168 & 0.050 & 0.931 \\ \hline
        \textit{coffee} & Ours & \textbf{0.266} & \textbf{0.093} & \textbf{0.928} \\
        \textit{martini} & CTRL-D & 0.252 & 0.081 & 0.917 \\ \hline
        \textit{mochi-} & Ours & \textbf{0.288} & \textbf{0.329} & \textbf{0.950} \\
        \textit{high-five} & CTRL-D & 0.278 & 0.310 & 0.945 \\ \hline
        \textit{sriracha-} & Ours & \textbf{0.245} & \textbf{0.219} & \textbf{0.905} \\
        \textit{tree} & CTRL-D & 0.210 & 0.208 & 0.900 \\ \hline
    \end{tabular}
    \vspace{-2mm}
    \caption{\textbf{Quantiative Comparison.} We provide a quantitaive comparison with CTRL-D on the ``\textit{pillow}'', ``\textit{mochi-high-five}'', and ``\textit{sriracha-tree}'' scenes from DyCheck dataset and ``\textit{coffee\_martini}'' from N3DV dataset. The evaluation is conducted using three metrics: CLIP text-image similarity ($\mathrm{CLIP_S}$), CLIP text-image direction similarity ($\mathrm{CLIP_D}$), and Consistency~\cite{huang2024vbench} metrics. Bold numbers indicate the best performance for each metric. More editing details of these scenes are stated in the supplementary material.}
    \vspace{-7mm}
    \label{tab:quantiative}
\end{table}

\noindent\textbf{Quantitative Comparison.}
In the qualitative analysis, both our method and the baseline CTRL-D demonstrate strong instruction-following capabilities. To further validate the effectiveness of our approach, we conduct quantitative comparisons between \method and CTRL-D.
We employ several widely used metrics for 4D editing evaluation: a subject consistency score (denote as \textbf{Consistency}) from VBench~\cite{huang2024vbench} to assess consistency of the edited objects across views and time, CLIP text-image similarity (denote as $\mathbf{CLIP_S}$) to measure alignment between the edited images and the prompt semantics, and CLIP text-image direction similarity (denotes as $\mathbf{CLIP_D}$) to evaluate the semantic change aligns with the visual change. The detailed results are shown in Table~\ref{tab:quantiative}.
As observed, our method consistently outperforms CTRL-D across all metrics in both test scenes, including ``pillow'' from the DyCheck dataset and ``coffee\_martini'' from the N3DV dataset. These results not only confirm the superior consistency of our edits but also highlight the stronger instruction adherence achieved by \method.

\begin{figure}[t!]
    \centering
    \includegraphics[width=\linewidth]{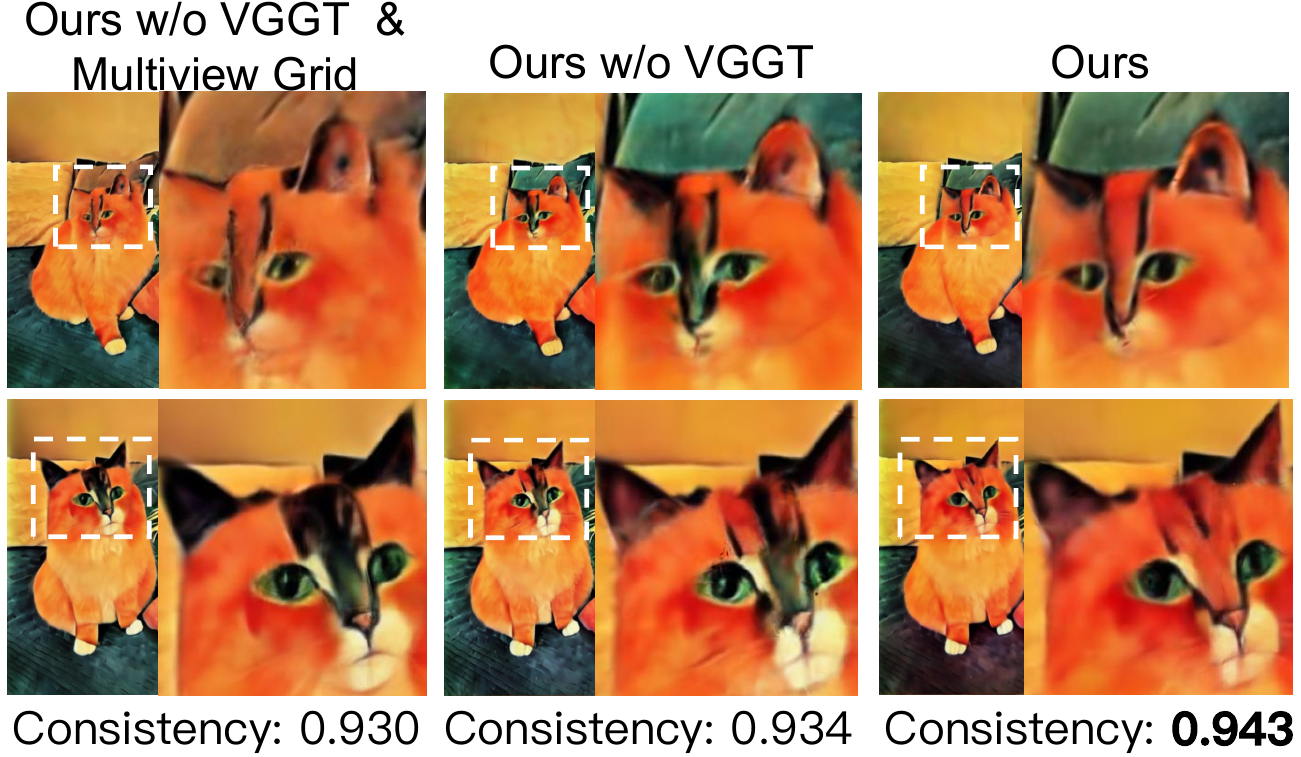}
    \vspace{-5mm}
    \caption{Ablation study of VGGT features and multi-view grid on the ``\textit{mochi-high-five}'' scene from DyCheck dataset. Our complete method achieves more consistent results.}
    \vspace{-5mm}
    \label{fig:ab_vggt}
\end{figure}

\begin{figure*}
    \centering
    \includegraphics[width=\linewidth]{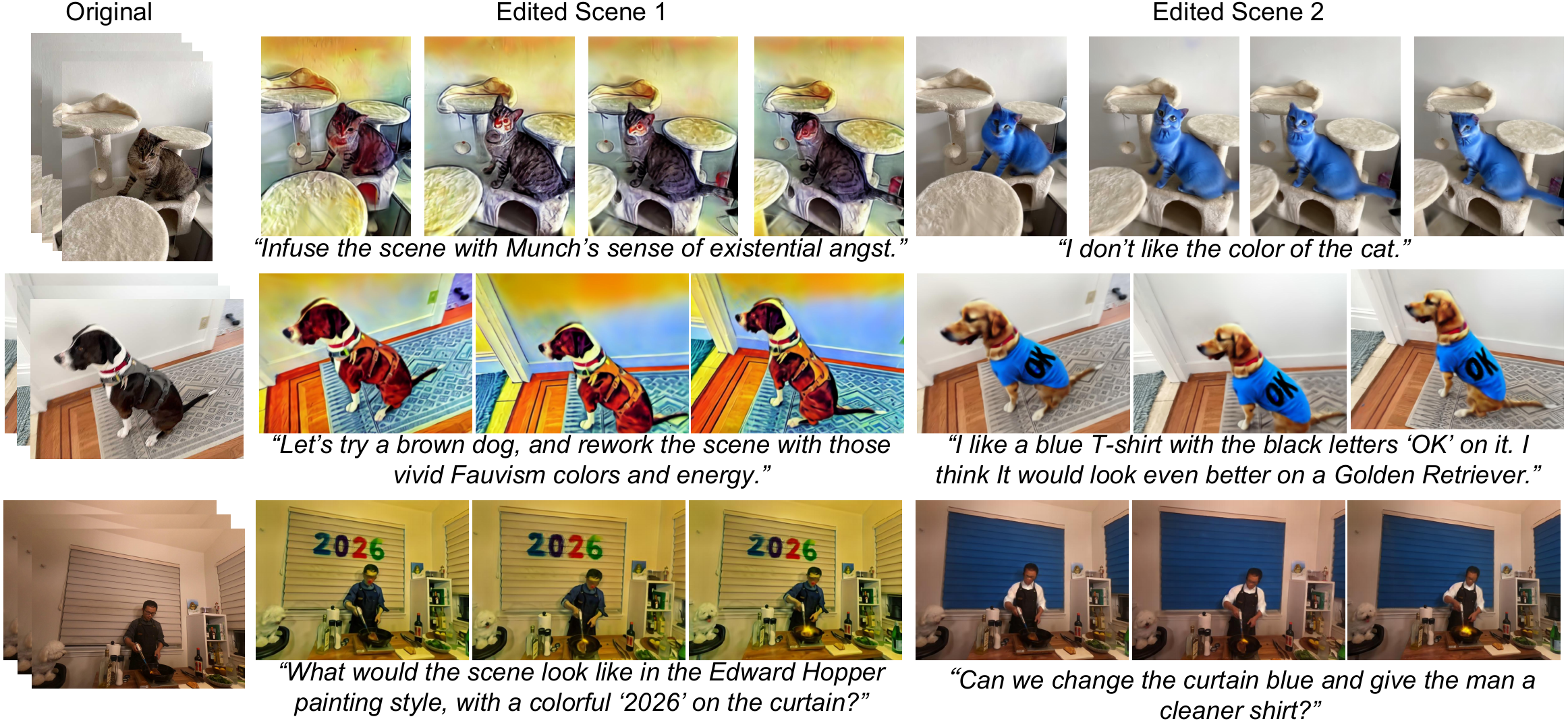}
    \vspace{-6mm}
    \caption{Visualization results on the ``\textit{srirahca-tree}'' and ``\textit{haru-sit}'' scenes from the monocular DyCheck dataset, and the ``\textit{flame\_steak}'' scene from the multi-camera N3DV dataset, under various editing prompts.}
    \vspace{-3mm}
    \label{fig:visulization}
\end{figure*}

\subsection{4.3 Ablation Studies}
\noindent\textbf{Ablation of our 4D-aware IP2P.} We conduct ablation studies on our 4D-aware IP2P to evaluate the impact of two key components, including the VGGT features and the multi-view grid. The experiments are performed on the ``mochi-high-fox'' scene from the DyCheck dataset. Specifically, we examine three variants: (1) \textbf{Ours}, which is our \method; (2) and \textbf{Ours w/o VGGT}, which removes the VGGT features from the 4D-aware IP2P; and (3) \textbf{Ours w/o VGGT \& Multiview Grid}, which removes both VGGT features and multi-view grid from our 4D-aware IP2P. As shown in Figure~\ref{fig:ab_vggt}, our method achieves the highest consistency across both viewpoint and time, particularly in the forehead region of the cat. In contrast, the variant with VGGT features \& multi-view grid shows the worst performance in terms of consistency, with noticeable artifacts in the same region.
Furthermore, our full model also outperforms all baselines on VBench, a quantitative metric for evaluating temporal and view consistency. These results demonstrate that incorporating both VGGT features and multi-view grid into the 4D-aware IP2P is crucial for achieving coherent editing.

\begin{table}[t!]
    \centering
    \setlength{\tabcolsep}{2.8pt}
    \begin{tabular}{c|c|c|c|c}
    \hline
         Method & PSNR $\uparrow$ & SSIM $\uparrow$ & LPIPS $\downarrow$ & Consistency $\uparrow$ \\ \hline
         Ours & \textbf{40.1} & \textbf{0.967} & \textbf{0.077} & \textbf{0.950} \\
         Ours w/o GSE & 35.0 & 0.935 & 0.141 & 0.944 \\ \hline
    \end{tabular}
    \vspace{-2mm}
    \caption{Ablation study of our Gaussian Selector (GSE) with quantitative analysis.
We report PSNR, SSIM, LPIPS, and consistency metrics on the non-edited regions.
Bold values indicate the best performance for each metric.}
    \label{tab:ab_mask}
    \vspace{-6mm}
\end{table}

\noindent\textbf{Ablation of our Gaussian Selection.} We also conduct an ablation study on the Gaussian selection technique, as shown in Figure~\ref{fig:ab_mask_LIUM} (a). Our full \method preserves sharp textural details in non-edited regions. In contrast, the absence of the Gaussian Selection leads to continuous updates of non-edited Gaussians during training, ultimately causing background artifacts and loss of sharpness. We also analyze the quantitative metrics, including PSNR, SSIM, LPIPS, and consistency, within the non-editing regions, as shown in Table~\ref{tab:ab_mask}. Our complete \method achieves superior performance across all metrics.
This observation highlights the critical role of the Gaussian Selector in maintaining the consistency of non-edited regions throughout the 4DGS optimization process.

\noindent\textbf{Ablation of our LLM for User Understanding.} We further conduct an ablation study on the ``mochi-high-five'' scene from the DyCheck dataset to evaluate the effectiveness of the LLM-based user understanding module, as shown in Figure~\ref{fig:ab_mask_LIUM} (b). With this module, the prompt ``\textit{Turn the cat to a fox, and then give the fox a pair of sunglasses}'' is split into two atomic tasks, including the first task-``\textit{Turn the cat to a fox}'', and the second task-``\textit{Give the fox a pair of sunglasses}''. Edited the 4DGS with these two atomic tasks leads to significantly better results than using the original compound prompt, especially in the execution of the second atomic task. The results demonstrate that our LLM-based intent understanding module plays an important role in improving the consistency and accuracy of complex 4DGS editing.

\subsection{4.4 Visualization} We design a series of prompts to edit 4DGS across various scenes, including ``sriracha-tree'' and ``haru-sit'' from the DyCheck dataset, and ``flame\_steak'' from the N3DV dataset. For each input prompt, the resulting edited 4DGS are visualized in Figure~\ref{fig:visulization}. Our method demonstrates strong instruction-following capability and editing consistency across diverse prompts and scenes. These results highlight the adaptability of our approach to various prompt types and scene configurations, showcasing the robustness of \method in real-world 4D editing scenarios. The details of each atomic task and more visualization results can be found in the supplementary material.

\section{5. Conclusion}

In this paper, we propose \method, an interactive 4DGS editing framework that ensures both spatial and temporal consistency. Our approach introduces a LLM for user intent understanding and task planning to allow diverse instructions, and designs a 4D-aware IP2P network that incorporates VGGT features and multi-view inputs to achieve consistent editing across views and time. We envision our method handling a wide range of user instructions, including highly abstract and complex commands, while significantly improving consistency compared to related works and producing visually pleasing results. We further evaluate our method using various metrics, demonstrating that it outperforms baselines and highlighting its superior performance.
We expect this framework to open new possibilities for future interactive 4D scene editing applications, empowering more intuitive and consistent 4D content creation.

\bibliography{aaai2026}

\begin{thebibliography}{32}
\providecommand{\natexlab}[1]{#1}

\bibitem[{Brooks, Holynski, and Efros(2023)}]{brooks2023instructpix2pix}
Brooks, T.; Holynski, A.; and Efros, A.~A. 2023.
\newblock Instructpix2pix: Learning to follow image editing instructions.
\newblock In \emph{Proceedings of the IEEE/CVF conference on computer vision and pattern recognition}, 18392--18402.

\bibitem[{Chen, Lyu, and Wang(2023)}]{chen2023neuraleditor}
Chen, J.-K.; Lyu, J.; and Wang, Y.-X. 2023.
\newblock Neuraleditor: Editing neural radiance fields via manipulating point clouds.
\newblock In \emph{Proceedings of the IEEE/CVF conference on computer vision and pattern recognition}, 12439--12448.

\bibitem[{Gao et~al.(2022)Gao, Li, Tulsiani, Russell, and Kanazawa}]{dycheck}
Gao, H.; Li, R.; Tulsiani, S.; Russell, B.; and Kanazawa, A. 2022.
\newblock Monocular dynamic view synthesis: A reality check.
\newblock \emph{Advances in Neural Information Processing Systems}, 35: 33768--33780.

\bibitem[{Gao et~al.(2025)Gao, Li, Zhuang, Zhang, Hu, Zhang, Yao, Shan, and Quan}]{gao2025mani}
Gao, X.; Li, X.; Zhuang, Y.; Zhang, Q.; Hu, W.; Zhang, C.; Yao, Y.; Shan, Y.; and Quan, L. 2025.
\newblock Mani-gs: Gaussian splatting manipulation with triangular mesh.
\newblock In \emph{Proceedings of the Computer Vision and Pattern Recognition Conference}, 21392--21402.

\bibitem[{Haque et~al.(2023)Haque, Tancik, Efros, Holynski, and Kanazawa}]{haque2023instruct}
Haque, A.; Tancik, M.; Efros, A.~A.; Holynski, A.; and Kanazawa, A. 2023.
\newblock Instruct-nerf2nerf: Editing 3d scenes with instructions.
\newblock In \emph{Proceedings of the IEEE/CVF international conference on computer vision}, 19740--19750.

\bibitem[{He, Wu, and Gilitschenski(2025)}]{he2025ctrl}
He, K.; Wu, C.-H.; and Gilitschenski, I. 2025.
\newblock CTRL-D: Controllable Dynamic 3D Scene Editing with Personalized 2D Diffusion.
\newblock In \emph{Proceedings of the Computer Vision and Pattern Recognition Conference}, 26630--26640.

\bibitem[{Huang et~al.(2024)Huang, He, Yu, Zhang, Si, Jiang, Zhang, Wu, Jin, Chanpaisit et~al.}]{huang2024vbench}
Huang, Z.; He, Y.; Yu, J.; Zhang, F.; Si, C.; Jiang, Y.; Zhang, Y.; Wu, T.; Jin, Q.; Chanpaisit, N.; et~al. 2024.
\newblock Vbench: Comprehensive benchmark suite for video generative models.
\newblock In \emph{Proceedings of the IEEE/CVF Conference on Computer Vision and Pattern Recognition}, 21807--21818.

\bibitem[{Jambon et~al.(2023)Jambon, Kerbl, Kopanas, Diolatzis, Leimk{\"u}hler, and Drettakis}]{jambon2023nerfshop}
Jambon, C.; Kerbl, B.; Kopanas, G.; Diolatzis, S.; Leimk{\"u}hler, T.; and Drettakis, G. 2023.
\newblock Nerfshop: Interactive editing of neural radiance fields.
\newblock \emph{Proceedings of the ACM on Computer Graphics and Interactive Techniques}, 6(1).

\bibitem[{Jang, Gu, and Poole(2016)}]{jang2016categorical}
Jang, E.; Gu, S.; and Poole, B. 2016.
\newblock Categorical reparameterization with gumbel-softmax.
\newblock \emph{arXiv preprint arXiv:1611.01144}.

\bibitem[{Jiang et~al.(2024)Jiang, Yu, Xie, Li, Feng, Wang, Li, Lau, Gao, Yang et~al.}]{jiang2024vr}
Jiang, Y.; Yu, C.; Xie, T.; Li, X.; Feng, Y.; Wang, H.; Li, M.; Lau, H.; Gao, F.; Yang, Y.; et~al. 2024.
\newblock Vr-gs: A physical dynamics-aware interactive gaussian splatting system in virtual reality.
\newblock In \emph{ACM SIGGRAPH 2024 Conference Papers}, 1--1.

\bibitem[{Kerbl et~al.(2023)Kerbl, Kopanas, Leimk{\"u}hler, and Drettakis}]{kerbl20233d}
Kerbl, B.; Kopanas, G.; Leimk{\"u}hler, T.; and Drettakis, G. 2023.
\newblock 3D Gaussian splatting for real-time radiance field rendering.
\newblock \emph{ACM Trans. Graph.}, 42(4): 139--1.

\bibitem[{Kwon, Cho, and Kim(2025)}]{kwon2025efficient}
Kwon, J.; Cho, H.; and Kim, J. 2025.
\newblock Efficient Dynamic Scene Editing via 4D Gaussian-based Static-Dynamic Separation.
\newblock In \emph{Proceedings of the Computer Vision and Pattern Recognition Conference}, 26855--26865.

\bibitem[{Li et~al.(2022)Li, Slavcheva, Zollhoefer, Green, Lassner, Kim, Schmidt, Lovegrove, Goesele, Newcombe et~al.}]{N3DV}
Li, T.; Slavcheva, M.; Zollhoefer, M.; Green, S.; Lassner, C.; Kim, C.; Schmidt, T.; Lovegrove, S.; Goesele, M.; Newcombe, R.; et~al. 2022.
\newblock Neural 3d video synthesis from multi-view video.
\newblock In \emph{Proceedings of the IEEE/CVF conference on computer vision and pattern recognition}, 5521--5531.

\bibitem[{Liu et~al.(2024)Liu, Zeng, Ren, Li, Zhang, Yang, Jiang, Li, Yang, Su et~al.}]{liu2024grounding}
Liu, S.; Zeng, Z.; Ren, T.; Li, F.; Zhang, H.; Yang, J.; Jiang, Q.; Li, C.; Yang, J.; Su, H.; et~al. 2024.
\newblock Grounding dino: Marrying dino with grounded pre-training for open-set object detection.
\newblock In \emph{European conference on computer vision}, 38--55. Springer.

\bibitem[{Mildenhall et~al.(2021)Mildenhall, Srinivasan, Tancik, Barron, Ramamoorthi, and Ng}]{mildenhall2021nerf}
Mildenhall, B.; Srinivasan, P.~P.; Tancik, M.; Barron, J.~T.; Ramamoorthi, R.; and Ng, R. 2021.
\newblock Nerf: Representing scenes as neural radiance fields for view synthesis.
\newblock \emph{Communications of the ACM}, 65(1): 99--106.

\bibitem[{Mou, Chen, and Wang(2024)}]{mou2024instruct}
Mou, L.; Chen, J.-K.; and Wang, Y.-X. 2024.
\newblock Instruct 4d-to-4d: Editing 4d scenes as pseudo-3d scenes using 2d diffusion.
\newblock In \emph{Proceedings of the IEEE/CVF Conference on Computer Vision and Pattern Recognition}, 20176--20185.

\bibitem[{Park et~al.(2021)Park, Sinha, Barron, Bouaziz, Goldman, Seitz, and Martin-Brualla}]{park2021nerfies}
Park, K.; Sinha, U.; Barron, J.~T.; Bouaziz, S.; Goldman, D.~B.; Seitz, S.~M.; and Martin-Brualla, R. 2021.
\newblock Nerfies: Deformable neural radiance fields.
\newblock In \emph{Proceedings of the IEEE/CVF international conference on computer vision}, 5865--5874.

\bibitem[{Peng et~al.(2022)Peng, Yan, Liu, Cheng, Guan, Pan, Zhai, and Yang}]{peng2022cagenerf}
Peng, Y.; Yan, Y.; Liu, S.; Cheng, Y.; Guan, S.; Pan, B.; Zhai, G.; and Yang, X. 2022.
\newblock Cagenerf: Cage-based neural radiance field for generalized 3d deformation and animation.
\newblock \emph{Advances in Neural Information Processing Systems}, 35: 31402--31415.

\bibitem[{Poole et~al.()Poole, Jain, Barron, and Mildenhall}]{pooledreamfusion}
Poole, B.; Jain, A.; Barron, J.~T.; and Mildenhall, B. ????
\newblock DreamFusion: Text-to-3D using 2D Diffusion.
\newblock In \emph{The Eleventh International Conference on Learning Representations}.

\bibitem[{Pumarola et~al.(2021)Pumarola, Corona, Pons-Moll, and Moreno-Noguer}]{pumarola2021d}
Pumarola, A.; Corona, E.; Pons-Moll, G.; and Moreno-Noguer, F. 2021.
\newblock D-nerf: Neural radiance fields for dynamic scenes.
\newblock In \emph{Proceedings of the IEEE/CVF conference on computer vision and pattern recognition}, 10318--10327.

\bibitem[{Shao et~al.(2024)Shao, Sun, Peng, Zheng, Zhou, Zhang, and Liu}]{shao2024control4d}
Shao, R.; Sun, J.; Peng, C.; Zheng, Z.; Zhou, B.; Zhang, H.; and Liu, Y. 2024.
\newblock Control4d: Efficient 4d portrait editing with text.
\newblock In \emph{Proceedings of the IEEE/CVF Conference on Computer Vision and Pattern Recognition}, 4556--4567.

\bibitem[{Wang et~al.(2022)Wang, Chai, He, Chen, and Liao}]{wang2022clip}
Wang, C.; Chai, M.; He, M.; Chen, D.; and Liao, J. 2022.
\newblock Clip-nerf: Text-and-image driven manipulation of neural radiance fields.
\newblock In \emph{Proceedings of the IEEE/CVF conference on computer vision and pattern recognition}, 3835--3844.

\bibitem[{Wang et~al.(2023{\natexlab{a}})Wang, He, Chai, Chen, and Liao}]{wang2023mesh}
Wang, C.; He, M.; Chai, M.; Chen, D.; and Liao, J. 2023{\natexlab{a}}.
\newblock Mesh-guided neural implicit field editing.
\newblock \emph{arXiv preprint arXiv:2312.02157}.

\bibitem[{Wang et~al.(2023{\natexlab{b}})Wang, Jiang, Chai, He, Chen, and Liao}]{wang2023nerf}
Wang, C.; Jiang, R.; Chai, M.; He, M.; Chen, D.; and Liao, J. 2023{\natexlab{b}}.
\newblock Nerf-art: Text-driven neural radiance fields stylization.
\newblock \emph{IEEE Transactions on Visualization and Computer Graphics}, 30(8): 4983--4996.

\bibitem[{Wang et~al.(2024)Wang, Zhong, Chai, He, Chen, and Liao}]{wang2024chat2layout}
Wang, C.; Zhong, H.; Chai, M.; He, M.; Chen, D.; and Liao, J. 2024.
\newblock Chat2Layout: Interactive 3D furniture layout with a multimodal LLM.
\newblock \emph{arXiv preprint arXiv:2407.21333}.

\bibitem[{Wang et~al.(2025)Wang, Chen, Karaev, Vedaldi, Rupprecht, and Novotny}]{wang2025vggt}
Wang, J.; Chen, M.; Karaev, N.; Vedaldi, A.; Rupprecht, C.; and Novotny, D. 2025.
\newblock Vggt: Visual geometry grounded transformer.
\newblock In \emph{Proceedings of the Computer Vision and Pattern Recognition Conference}, 5294--5306.

\bibitem[{Wang et~al.(2023{\natexlab{c}})Wang, Lu, Wang, Bao, Li, Su, and Zhu}]{wang2023prolificdreamer}
Wang, Z.; Lu, C.; Wang, Y.; Bao, F.; Li, C.; Su, H.; and Zhu, J. 2023{\natexlab{c}}.
\newblock Prolificdreamer: High-fidelity and diverse text-to-3d generation with variational score distillation.
\newblock \emph{Advances in neural information processing systems}, 36: 8406--8441.

\bibitem[{Wu et~al.(2024{\natexlab{a}})Wu, Yi, Fang, Xie, Zhang, Wei, Liu, Tian, and Wang}]{wu20244d}
Wu, G.; Yi, T.; Fang, J.; Xie, L.; Zhang, X.; Wei, W.; Liu, W.; Tian, Q.; and Wang, X. 2024{\natexlab{a}}.
\newblock 4d gaussian splatting for real-time dynamic scene rendering.
\newblock In \emph{Proceedings of the IEEE/CVF conference on computer vision and pattern recognition}, 20310--20320.

\bibitem[{Wu et~al.(2024{\natexlab{b}})Wu, Yi, Fang, Xie, Zhang, Wei, Liu, Tian, and Wang}]{4dgs}
Wu, G.; Yi, T.; Fang, J.; Xie, L.; Zhang, X.; Wei, W.; Liu, W.; Tian, Q.; and Wang, X. 2024{\natexlab{b}}.
\newblock 4d gaussian splatting for real-time dynamic scene rendering.
\newblock In \emph{Proceedings of the IEEE/CVF conference on computer vision and pattern recognition}, 20310--20320.

\bibitem[{Wu et~al.(2023)Wu, Ge, Wang, Lei, Gu, Shi, Hsu, Shan, Qie, and Shou}]{wu2023tune}
Wu, J.~Z.; Ge, Y.; Wang, X.; Lei, S.~W.; Gu, Y.; Shi, Y.; Hsu, W.; Shan, Y.; Qie, X.; and Shou, M.~Z. 2023.
\newblock Tune-a-video: One-shot tuning of image diffusion models for text-to-video generation.
\newblock In \emph{Proceedings of the IEEE/CVF international conference on computer vision}, 7623--7633.

\bibitem[{Yang et~al.(2023)Yang, Yang, Pan, and Zhang}]{yangreal}
Yang, Z.; Yang, H.; Pan, Z.; and Zhang, L. 2023.
\newblock Real-time Photorealistic Dynamic Scene Representation and Rendering with 4D Gaussian Splatting.
\newblock In \emph{The Twelfth International Conference on Learning Representations}.

\bibitem[{Zhao, Lee, and Hsu(2023)}]{zhao2023large}
Zhao, Z.; Lee, W.~S.; and Hsu, D. 2023.
\newblock Large language models as commonsense knowledge for large-scale task planning.
\newblock \emph{Advances in neural information processing systems}, 36: 31967--31987.

\end{thebibliography}


\end{document}